# Metrics for Markov Decision Processes with Infinite State Spaces

**Norm Ferns**
School of Computer Science
McGill University
Montréal, Canada, H3A 2A7
nferns@cs.mcgill.ca

**Prakash Panangaden**
School of Computer Science
McGill University
Montréal, Canada, H3A 2A7
prakash@cs.mcgill.ca

**Doina Precup**
School of Computer Science
McGill University
Montréal, Canada, H3A 2A7
dprecup@cs.mcgill.ca

## Abstract

We present metrics for measuring state similarity in Markov decision processes (MDPs) with infinitely many states, including MDPs with continuous state spaces. Such metrics provide a stable quantitative analogue of the notion of bisimulation for MDPs, and are suitable for use in MDP approximation. We show that the optimal value function associated with a discounted infinite horizon planning task varies continuously with respect to our metric distances.

## 1 Introduction

Markov decision processes (MDPs) offer a popular mathematical tool for planning and learning in the presence of uncertainty (Boutilier et al., 1999). MDPs are a standard formalism for describing multi-stage decision making in probabilistic environments. The objective of the decision making is to maximize a cumulative measure of long-term performance, called the *return*. Dynamic programming algorithms, e.g., value iteration or policy iteration (Puterman, 1994), allow us to compute the optimal expected return for any state, as well as the way of behaving (policy) that generates this return. However, in many practical applications, the state space of an MDP is simply too large, possibly infinite or even continuous, for such standard algorithms to be applied. A typical means of overcoming such circumstances is to partition the state space in the hope of obtaining an "essentially equivalent" reduced system. One defines a new MDP over the partition blocks, and if it is small enough, it can be solved by classical methods. The hope is that optimal values and policies for the reduced MDP can be extended to optimal values and policies for the original MDP.

Recent MDP research on defining equivalence relations on MDPs (Givan et al., 2003) has built on the notion of strong probabilistic bisimulation from concurrency theory. Probabilistic bisimulation was introduced by Larsen and Skou (1991) based on bisimulation for nondeterministic (nonprobabilistic) systems due to Park (1981) and Milner (1980). Henceforth when we say "bisimulation" we will mean strong probabilistic bisimulation.

In a probabilistic setting, bisimulation can be described as an equivalence relation that relates two states precisely when they have the same probability of transitioning to classes of equivalent states. The extension of bisimulation to transition systems with rewards was carried out in the context of MDPs by Givan, Dean and Greig (2003) and in the context of performance evaluation by Bernardo and Bravetti (2003). In both cases, the motivation is to use the equivalence relation to aggregate the states and get smaller state spaces. The basic notion of bisimulation is modified only slightly by the introduction of rewards.

However, it has been well established for a while now that use of exact equivalences in quantitative systems is problematic. A notion of equivalence is two-valued: two states are either equivalent or not equivalent. A small perturbation of the transition probabilities can make two equivalent states no longer equivalent. In short, any kind of equivalence is too unstable to perturbations of the numerical values of the transition probabilities. A natural remedy is to use metrics. Metrics are natural quantitative analogues of the notion of equivalence relation: for example the triangle inequality is a natural quantitative analogue of transitivity. The metrics on which we focus here specify the degree to which objects of interest behave similarly.

Much of this work has been done in a very general setting, using the labelled Markov process (LMP) model (Blute et al., 1997; Desharnais et al., 2002a). Previous metrics (Desharnais et al., 1999; van Breugel & Worrell, 2001; Desharnais et al., 2002b) (more precisely pseudometrics or semimetrics) have quantitatively generalized bisimulation by assigning distance zero to states that are bisimilar, distance one to states that are easily distinguishable, and an intermediate distance to those in between.

Van Breugel and Worrell (2001) showed how, in a simplified setting of finite state space LMPs, metric distances could be calculated in polynomial time. This work, along with that of others (Desharnais et al., 2002b), was then adapted to finite MDPs (Ferns et al., 2004). The current authors used fixed point theory to construct metrics, each of which had bisimulation as its kernel, was sensitive to perturbations in MDP parameters, and provided bounds on the optimal values of states. We showed how to compute the metrics up to any prescribed degree of accuracy and then used them to directly aggregate sample finite MDPs.

In this paper, we present a significant generalization of these previous results to MDPs with continuous state spaces. The linear programming arguments we used in our previous work no longer apply, and we have to use measure theory and duality theory on continuous state spaces. The mathematical theory is interesting in its own right. Although continuous MDPs are of great interest for practical applications, e.g. in the areas of automated control and robotics, the existing methods for measuring distances between states, for the purpose of state aggregation as well as other approximation methods are still largely heuristic. As a result, it is hard to provide guaranteed error bounds between the correct and the approximate value function. It is also difficult to determine the impact that structural changes in the approximator would have on the quality on the approximation. The metrics we define in this paper allow the definition of error bounds for value functions. These bounds can be used as a tool in the analysis of existing approximation schemes.

The paper is organized as follows. In sections 2 and 3 we provide the theoretical tools necessary for the construction of our metrics. The actual construction is carried out in section 4, where we also argue that our metrics are the most suitably stable tools for analyzing MDP state space compression. Section 5 provides a proof of value function continuity with respect to our metrics. In section 6 we provide a simple illustration of metric use in approximation. Finally, section 7 contains our conclusions and directions for future work.

## 2 Background

### 2.1 Markov Decision Processes

Let $(S, A, P, r)$ be a Markov decision process (MDP), where $S$ is complete separable metric space equipped with its Borel sigma algebra $\Sigma$, $A$ is a finite set of actions, $r : S \times A \to \mathbb{R}$ is a measurable reward function, and $P : S \times A \times \Sigma \to [0, 1]$ is a labeled stochastic transition kernel, i.e.

- $\forall a \in A, \forall s \in S,\ P(s, a, \cdot) : \Sigma \to [0, 1]$ is a probability measure, and
- $\forall a \in A, \forall X \in \Sigma,\ P(\cdot, a, X) : S \to [0, 1]$ is a measurable function.

We will use the following notation: for $a \in A$ and $s \in S$, $P_s^a$ denotes $P(s, a, \cdot)$ and $r_s^a$ denotes $r(s, a)$. Given measure $P$ and integrable function $f$, we denote the integral of $f$ with respect to $P$ by $P(f)$.

We also make the following assumptions:

1. $B := \sup_{s,s',a} |r_s^a - r_{s'}^a| < \infty$.

2. For each $a \in A$, $r(\cdot, a)$ is continuous on $S$.

3. For each $a \in A$, $P_s^a$ is (weakly) continuous as a function of $s$, i.e. if $s_n$ tends to $s$ in $S$ then for every bounded continuous function $f : S \to \mathbb{R}$, $P_{s_n}^a(f)$ tends to $P_s^a(f)$.

The first assumption is a direct consequence of the standard assumption that rewards are bounded. The second assumption is non-standard, but very mild. In general, rewards in an MDP are not assumed to vary continuously (e.g., in goal-directed tasks). However, it is generally assumed that there would be a finite or countable number of discontinuities. In this case, it is easy to transform the reward structure into one that is continuous and arbitrarily close to the original one, e.g. by applying smoothing sigmoid functions at the points of discontinuity. The third assumption is a continuity assumption on the transition probabilities, and satisfied by most "reasonable" systems (including physical systems of interest in control and robotics).

The discounted, infinite horizon planning task in an MDP is to determine a policy $\pi : S \to A$ that maximizes the value of every state, $V^\pi(s) = E[\sum_{t=0}^\infty r_t | s_0 = s, \pi]$, where $s_0$ is the state at time 0, $r_t$ is the reward achieved at time $t$, $\gamma$ is a discount factor in $(0, 1)$, and the expectation is taken by following the state dynamics induced by $\pi$. The function $V^\pi$ is called the value function of policy $\pi$. The optimal value function $V^*$, associated with an optimal policy, is the unique solution of the fixed point equation

$$V^*(s) = \max_{a \in A}(r_s^a + \gamma P_s^a(V^*))$$

and can be used to directly determine an optimal policy, provided it is computable. Note that in general the optimal value function need not be measurable, in which case the fixed point equation would be invalid.

However, under assumptions 1-3, this cannot be the case (Puterman, 1994, theorem 6.2.12). In fact, in this case, the optimal value function can be computed as the limit of a sequence of iterates. Define $V_0 = 0$ and $V_{n+1}(s) = \max_{a \in A}(r_s^a + \gamma P_s^a(V_n))$. Then the $V_n$s converge to $V^*$ in the uniform (max-norm) metric.

Of course, for this computation to work in practice it would be desirable to work with a small discretized version of the given MDP. This brings about the problem of approximation, and finding a formal definition which characterizes when states are equivalent (and hence can be lumped together). The correct equivalence relation is bisimulation.

## 2.2 Bisimulation

Bisimulation is a notion of behavioural equivalence, the strongest of a plethora of equivalence relations considered in concurrency theory. Bisimulation can be defined solely in terms of relations or using fixed point theory (so called co-induction). The latter will be useful for our purposes, but first requires some basic definitions and tools from fixed point theory on lattices that can be found in any basic text (Winskel, 1993).

Let $(L, \preceq)$ be a partial order. If it has least upper bounds and greatest lower bounds of arbitrary subsets of elements, then it is said to be a *complete lattice*. A function $f : L \to L$ is said to be *monotone* if $x \preceq x'$ implies $f(x) \preceq f(x')$. A point $x$ in $L$ is said to be a *prefixed point* if $f(x) \preceq x$, a *postfixed point* if $x \preceq f(x)$ and a *fixed point* if $x = f(x)$. The importance of these definitions arises in the following theorem.

**Theorem 2.1.** [1] *Let $L$ be a complete lattice, and suppose $f : L \to L$ is monotone. Then $f$ has a least fixed point, which is also its least prefixed point, and $f$ has a greatest fixed point, which is also its greatest postfixed point.*

Let $REL$ be the complete lattice of binary relations on $S$ with the usual subset ordering. We say a set in $X$ is $R$-closed if the collection of all those elements of $S$ that are reachable by $R$ from $X$ is itself contained in $X$. When $R$ is an equivalence relation this is equivalent to saying that $X$ is a union of $R$-equivalence classes. We write $R_{rst}$ for the reflexive, symmetric, transitive closure of $R$, and $\Sigma(R)$ for those $\Sigma$-measurable sets that are $R$-closed.

**Definition 2.2.** Define $\mathcal{F} : REL \to REL$ by $s\mathcal{F}(R)s' \Leftrightarrow \forall a \in A, \ r_s^a = r_{s'}^a$ and $\forall X \in \Sigma(R_{rst})$, $P_s^a(X) = P_{s'}^a(X)$. The greatest fixed point of $\mathcal{F}$ is *bisimulation*.

The existence of bisimulation is guaranteed by the fixed-point theorem. Unfortunately, as an exact equivalence, bisimulation suffers from issues of instability; that is, slight numerical differences in the MDP parameters, $r$ and $P$, can lead to vastly different bisimulation partitions. To get around this, one generalizes the notion of equivalence through metrics.

## 2.3 Metrics

**Definition 2.3.** A *semimetric*[2] on $S$ is a map $d : S \times S \to [0, \infty)$ such that for all $s$, $s'$, $s''$:

1. $s = s' \Rightarrow d(s, s') = 0$

2. $d(s, s') = d(s', s)$

3. $d(s, s'') \leq d(s, s') + d(s', s'')$

If the converse of the first axiom holds as well, we say $d$ is a *metric*. [3]

Recall that a function $h : S \times S \to \mathbb{R}$ is *lower semi-continuous* (lsc) if whenever $(s_n, s'_n)$ tends to $(s, s')$, $\liminf h(s_n, s'_n) \geq h(s, s')$. Here we are considering $S \times S$ to be endowed with the product topology. Note that lsc functions are product measurable.

Let $\mathcal{M}$ be the set of semimetrics on $S$ that are lsc on $S \times S$ and uniformly bounded, e.g. those assigning distance at most 1, and give it the usual pointwise ordering. Then $\mathcal{M}$ is a complete lattice. This follows because taking the pointwise supremum of an arbitrary collection of lsc functions yields a lsc function, and taking the pointwise supremum of an arbitrary collection of semimetrics yields a semimetric. Additionally, if we take $\mathcal{M}$ with the metric induced by the uniform norm, $\|h\| = \sup_{s,s'} |h(s, s')|$, then it is a complete metric space. The rich structure of $\mathcal{M}$ allows us to apply both the lattice theoretic fixed-point theorem and the more familiar Banach fixed-point theorem, provided we construct an appropriate map on $\mathcal{M}$.

Since bisimulation involves an exact matching of rewards and probabilistic transitions, the appropriate metric generalization should involve a metric on rewards and a metric on probability measures. The choice of reward metric is obvious: the usual Euclidean distance. The choice of probability metric, however, is not so obvious.

---

[1] This is an elementary theorem sometimes called the Knaester-Tarski theorem in the literature. In fact the Knaester-Tarski theorem is a much stronger statement to the effect that the collection of fixed points is itself a complete lattice.

[2] They are often called pseudometrics in the literature.

[3] For convenience we will use the terms metric and semimetric interchangeably; however, we really mean the latter.

# 3 Probability Metrics

There are numerous ways of defining a notion of distance between probability measures on a given space (Gibbs & Su, 2002). The particular probability semimetric of which we make use is known as the Kantorovich metric.

Given a semimetric $h \in \mathcal{M}$ and probability measures $P$ and $Q$ on $S$, the induced Kantorovich distance, $T_K(h)$, is defined by $T_K(h)(P,Q) = \sup_f (P(f) - Q(f))$, where the supremum is taken over all bounded measurable $f : S \to \mathbb{R}$ satisfying the Lipschitz condition: $f(x) - f(y) \leq h(x,y)$ for all $x, y \in S$. We write $Lip(h)$ for the set of all such functions.

In light of the definition of bisimulation, the importance of using the Kantorovich distance is made evident in the following lemma.

**Lemma 3.1.** *Let $h \in \mathcal{M}$. Then $T_K(h)(P,Q) = 0 \Leftrightarrow P(X) = Q(X)$, $\forall X \in \Sigma(Rel(h))$.*

*Proof.* $\Leftarrow$ Fix $\epsilon > 0$ and let $f \in Lip(h)$ such that $T_K(h)(P,Q) < P(f) - Q(f) + \epsilon$. WLOG $f \geq 0$. Choose $\psi$ a simple approximation (the usual one) to $f$ so that $T_K(h)(P,Q) < P(\psi) - Q(\psi) + 2\epsilon$. Let $\psi(S) = \{c_1, \ldots, c_k\}$ where the $c_i$ are distinct, $E_i = \psi^{-1}(\{c_i\})$, and $R = Rel(h)$. Then each $E_i$ is $R$-closed, for if $y \in R(E_i)$ then there is some $x \in E_i$ such that $h(x,y) = 0$. So $f(x) = f(y)$ and therefore, $\psi(x) = \psi(y)$. So $y \in E_i$. So by assumption $P(\psi) - Q(\psi) = \sum c_i P(E_i) - \sum c_i Q(E_i) = 0$. Thus, $T_K(h)(P,Q) = 0$.

$\Rightarrow$ Let $X \in \Sigma(R)$. Let $K \subseteq X$ be compact. Define $f(x) = \inf_{k \in K} h(x,k)$. Since a lsc function has a minimum on a compact set, we may write $f(x) = \min_{k \in K} h(x,k)$. In fact, $f$ is itself lsc (Puterman, 1994, theorem B.5). Since $f$ is measurable, $R(K) = f^{-1}(\{0\}) \in \Sigma(R)$. Now, since $P$ is tight (as $S$ is a complete separable metric space), $P(X) = \sup P(K)$ where the supremum is taken over all compact $K \subseteq X$. However, $K \subseteq X$ implies $K \subseteq R(K) \subseteq R(X) = X$. Since $R(K)$ is measurable, we have $P(X) = \sup P(R(K))$. Similarly, $Q(X) = \sup Q(R(K))$. Define $g_n = \max(0, 1 - nf)$. Then $g_n$ decreases to the indicator function on $R(K)$. Also, $g_n/n \in Lip(h)$, so by assumption $P(g_n/n) = Q(g_n/n)$. Multiplying by $n$ and taking limits gives $P(R(K)) = Q(R(K))$ and we are done. □

The Kantorovich metric arose in the study of optimal mass transportation. The following description is due to Villani (2002): assume we are given a pile of sand and a hole, occupying measurable spaces $(X, \Sigma_X)$ and $(Y, \Sigma_Y)$, each representing a copy of $(S, \Sigma)$. The pile of sand and the hole obviously have the same volume, and the mass of the pile is assumed to be normalized to 1. Let $P$ and $Q$ be measures on $X$ and $Y$ respectively, such that whenever $A \in \Sigma_X$ and $B \in \Sigma_Y$, $P[A]$ measures how much sand occupies $A$ and $Q[B]$ measures how much sand can be piled into $B$. Suppose further that we have some measurable cost function $h : X \times Y \to \mathbb{R}$, where $h(x,y)$ tells us how much it costs to transfer one unit of mass from a point $x \in X$ to a point $y \in Y$. Here we consider $h \in \mathcal{M}$. The goal is to determine a plan for transferring all the mass from $X$ to $Y$ while keeping the cost at a minimum. Such a transfer plan is modelled by a probability measure $\lambda$ on $(X \times Y, \Sigma_X \otimes \Sigma_Y)$, where $d\lambda(x,y)$ measures how much mass is transferred from location $x$ to $y$. Of course, for the plan to be valid we require that $\lambda[A \times Y] = P[A]$ and $\lambda[X \times B] = Q[B]$ for all measurable $A$ and $B$. A plan satisfying this condition is said to have marginals $P$ and $Q$, and we denote the collection of all such plans by $\Lambda(P,Q)$. We can now restate the goal formally as:

$$\text{minimize } h(\lambda) \text{ over } \lambda \in \Lambda(P,Q)$$

This is actually an instance of an infinite linear program. Fortunately, under very general circumstances, it has a solution and admits a dual formulation.

Let us first note that measures in $\Lambda(P,Q)$ can equivalently be characterized as those $\lambda$ satisfying:

$$P(\phi) + Q(\psi) = \lambda(\phi + \psi)$$

for all $(\phi, \psi) \in L^1(P) \times L^1(Q)$. As a consequence of this characterization we have the following inequality:

$$\sup_f (P(f) - Q(f)) \leq T_K(h)(P,Q) \leq \inf_{\lambda \in \Lambda(P,Q)} h(\lambda) \quad (1)$$

where $f$ is restricted to the continuous functions in $Lip(h)$.

The leftmost and rightmost terms in inequality (1) are examples of infinite linear programs in duality. It is a highly nontrivial result that there is no duality gap in this case, as a result of the Kantorovich-Rubinstein Duality Theorem with metric cost function (Rachev & Rüschendorf, 1998, theorems 4.15 & 4.28 and example 4.24; Villani, 2002).

In the case of a finite state space, this duality leads to a strongly polynomial time algorithm (in terms of the size of the state space) for calculating the Kantorovich metric (Orlin, 1988). Thus, one approach for calculating the Kantorovich metric is to discretize the linear program in some manner and solve a finite linear program (Rachev & Rüschendorf, 1998, section 5.3 with compact $S$). In further restricted settings, e.g. if $S$ is Euclidean and $h$ is continuous, more direct approximation schemes exist (Anderson & Nash, 1987, section 5.4). Issues of efficiency aside, the Kantorovich distance is computable.

We conclude this section by noting that if the state space metric is chosen to be the discrete metric, which assigns distance 1 to all pairs of unequal points, then the Kantorovich metric agrees with the total variation metric, defined as $d_{TV}(P,Q) = \sup_{X \in \Sigma} |P(X) - Q(X)|$. While simple to define, the total variation metric gives an overly strong measure of the numerical differences across probabilistic transitions to *all* measurable sets. Note, for example, that the distance between two point masses, $\delta_x$ and $\delta_y$, is always 1, unless $x = y$ exactly. Nevertheless, the total variation distance is commonly used in practice and can lead to interesting bounds.

## 4 Bisimulation Metrics

Our development of fixed point metrics mirrors closely the definition of bisimulation. In the following $c \in (0,1)$ is a discount factor, in the same vein as the discount factor $\gamma$ used in the definition and estimation of value functions. It determines the extent to which future transitions are taken into account when trying to distinguish states quantitatively. In section 2 we mentioned that $\mathcal{M}$ is a uniformly bounded set of lsc semimetrics. Here we fix that upper bound to be the constant $\alpha$ defined as $\frac{B}{1-c}$.

**Theorem 4.1.** *Let* $c \in (0,1)$. *Define* $F^c : \mathcal{M} \to \mathcal{M}$ *by*

$$F^c(h)(s,s') = \max_{a \in A}(|r_s^a - r_{s'}^a| + cT_K(h)(P_s^a, P_{s'}^a))$$

*Then* $F^c$ *has a least fixed point,* $d_{fix}^c$, *and* $Rel(d_{fix}^c)$ *is bisimulation.*

*Proof.* It is easy to see that $F^c$ is monotone on $\mathcal{M}$ and so existence of $d_{fix}^c$ follows from the Knaester-Tarski Theorem. It is important to note here that we are implicitly invoking the leftmost equality in (1) in order to correctly claim that the map taking $(s,s')$ to $T_K(h)(P_s^a, P_{s'}^a)$ is lsc.

By means of lemma 3.1 we find that for any $h$ in $\mathcal{M}$, $Rel(F^c(h)) = \mathcal{F}(Rel(h))$. Thus, $Rel(d_{fix}^c) = \mathcal{F}(Rel(d_{fix}^c))$ is a fixed point and so is contained in bisimulation. For the other direction, we consider the discrete bisimulation semimetric; note that it is not immediately clear that it is lsc. Call it $\mathbb{I}_{\not\sim}$. Let $l$ be the greatest lower bound in $\mathcal{M}$ of $\{\alpha \mathbb{I}_{\not\sim}\}$. Then $\sim \subseteq Rel(l)$. Thus, $\sim = \mathcal{F}(\sim) \subseteq \mathcal{F}(Rel(l)) = Rel(F^c(l))$, which implies $F^c(l) \leq \alpha \mathbb{I}_{\not\sim}$. Since $F^c(l) \in \mathcal{M}$, we must have $F^c(l) \leq l$. Since $d_{fix}^c$ is the least prefixed point of $F^c$, $d_{fix}^c \leq l \leq \alpha \mathbb{I}_{\not\sim}$, so that $\sim \subseteq Rel(d_{fix}^c)$. □

Thus, we have established existence of a metric that assigns distance zero to points exactly in the case when those points are bisimilar. Of course, $d_{fix}^c$ is not the only such metric; the discrete bisimulation metric, for example, is another. However, $d_{fix}^c$ is the most suitable candidate bisimulation metric for MDP analysis. Before we argue that this is the case, let us first note that $d_{fix}^c$ is in fact unique.

**Proposition 4.2.** *For any* $h_0 \in \mathcal{M}$,

$$\|d_{fix}^c - (F^c)^n(h_0)\| \leq \frac{c^n}{1-c}\|F^c(h_0) - h_0\|.$$

*In particular,* $\lim_n (F^c)^n(h_0) = d_{fix}^c$, *and* $d_{fix}^c$ *is the unique fixed point of* $F^c$.

*Proof.* This is simply an application of the Banach Fixed Point Theorem. Here we use the dual minimization form of $T_K(\cdot)$, as given in (1). Note that for all $h, h' \in \mathcal{M}$, and for all $s, s' \in S$,

$F^c(h)(s,s') - F^c(h')(s,s')$
$\leq c \max_{a \in A}(T_K(h)(P_s^a, P_{s'}^a) - T_K(h')(P_s^a, P_{s'}^a))$
$\leq c \max_{a \in A}(T_K(h - h' + h')(P_s^a, P_{s'}^a) - T_K(h')(P_s^a, P_{s'}^a))$
$\leq c \max_{a \in A}(T_K(\|h - h'\| + h')(P_s^a, P_{s'}^a) - T_K(h')(P_s^a, P_{s'}^a))$
$\leq c \max_{a \in A}(\|h - h'\| + T_K(h')(P_s^a, P_{s'}^a) - T_K(h')(P_s^a, P_{s'}^a))$
$\leq c\|h - h'\|$

Thus, $\|F^c(h) - F^c(h')\| \leq c\|h - h'\|$, so that $F^c$ is a contraction mapping and has an unique fixed point $d_{fix}^c$. □

As an immediate corollary of theorem 4.1 we find that bisimulation is a closed subset of $S \times S$, under the given restrictions on $r$ and $P$. So the discrete bisimulation metric, $\alpha \mathbb{I}_{\not\sim}$, is lsc, and in particular, $\{(F^c)^n(\alpha \mathbb{I}_{\not\sim})\}$ is a family of lsc semimetrics decreasing to $d_{fix}^c$, each of which has bisimulation as its kernel. The first iterate can be expressed in a more familiar form by noting that $T_K(\mathbb{I}_{\not\sim})(P,Q) = \sup_{X \in \Sigma(\sim)} |P(X) - Q(X)|$, which is the total variation distance of $P$ and $Q$ as defined over the fully compressed state space (see appendix for a proof). The advantage of using $d_{fix}^c$ over any of these iterates is that $d_{fix}^c$ is sensitive to perturbations in the MDP parameters. Formally, $d_{fix}^c$ is continuous in $r$ and $P$.

**Proposition 4.3.** *Suppose* $(r_i, P_i)$, $i = 1, 2$, *are MDP parameters, each satisfying the assumptions of section 2, and set* $B = \max(B_1, B_2)$. *Let* $d_1$ *and* $d_2$ *be the corresponding bisimulation metrics given by theorem 4.1 with discount factor c. Then*

$$\|d_1 - d_2\| \leq \frac{2}{1-c} \max_a \|r_1^a - r_2^a\|$$
$$+ \frac{2Bc}{(1-c)^2} \sup_{a,s} d_{TV}(P_{1,s}^a, P_{2,s}^a)$$

This result follows from the unwinding of the fixed point definitions of $d_1$ and $d_2$.

*Proof.* Since $Lip(\frac{d_2}{\|d_2\|}) \subseteq Lip(\mathbb{I}_{\neq})$, we first obtain the following inequality:

$$T_K(d_2)(P_{1,x}^a, P_{1,y}^a) - T_K(d_2)(P_{2,x}^a, P_{2,y}^a)$$
$$\leq \sup_{Lip(d_2)} (P_{1,x}^a(f) - P_{1,y}^a(f)) - (P_{2,x}^a(f) - P_{2,y}^a(f))$$
$$\leq \|d_2\| \sup_{Lip(\mathbb{I}_{\neq})} (P_{1,x}^a(\frac{f}{\|d_2\|}) - P_{2,x}^a(\frac{f}{\|d_2\|}))$$
$$- (P_{1,y}^a(\frac{f}{\|d_2\|}) - P_{2,y}^a(\frac{f}{\|d_2\|}))$$
$$\leq \|d_2\|(\sup_{Lip(\mathbb{I}_{\neq})} |P_{1,x}^a(g) - P_{2,x}^a(g)|$$
$$+ \sup_{Lip(\mathbb{I}_{\neq})} |P_{1,y}^a(g) - P_{2,y}^a(g)|)$$
$$\leq \|d_2\|(d_{TV}(P_{1,x}^a, P_{2,x}^a) + d_{TV}(P_{1,y}^a, P_{2,y}^a))$$

Here we are once more using the minimization form of $T_K$.

$$d_1(x,y) - d_2(x,y)$$
$$\leq \max_{a \in A}(|r_{1,x}^a - r_{1,y}^a| + cT_K(d_1)(P_{1,x}^a, P_{1,y}^a))$$
$$- \max_{a \in A}(|r_{2,x}^a - r_{2,y}^a| + cT_K(d_2)(P_{2,x}^a, P_{2,y}^a))$$
$$\leq \max_{a \in A}(|r_{1,x}^a - r_{1,y}^a| - |r_{2,x}^a - r_{2,y}^a|$$
$$+ c(T_K(d_1)(P_{1,x}^a, P_{1,y}^a) - T_K(d_2)(P_{2,x}^a, P_{2,y}^a)))$$
$$\leq \max_{a \in A}(|(r_{1,x}^a - r_{1,y}^a) - (r_{2,x}^a - r_{2,y}^a)|$$
$$+ c(T_K(d_1)(P_{1,x}^a, P_{1,y}^a) - T_K(d_2)(P_{1,x}^a, P_{1,y}^a))$$
$$+ c(T_K(d_2)(P_{1,x}^a, P_{1,y}^a) - T_K(d_2)(P_{2,x}^a, P_{2,y}^a))))$$
$$\leq \max_{a \in A}(|r_{1,x}^a - r_{2,x}^a| + |r_{1,y}^a - r_{2,y}^a| + c\|d_1 - d_2\|$$
$$+ 2c\|d_2\| \sup_s d_{TV}(P_{1,s}^a, P_{2,s}^a)))$$
$$\leq \max_{a \in A}(2\|r_1^a - r_2^a\| + c\|d_1 - d_2\|$$
$$+ 2c\|d_2\| \sup_s d_{TV}(P_{1,s}^a, P_{2,s}^a)))$$
$$\leq 2\max_{a \in A}\|r_1^a - r_2^a\| + c\|d_1 - d_2\|$$
$$+ 2c(\frac{B}{1-c}) \sup_{a,s} d_{TV}(P_{1,s}^a, P_{2,s}^a)))$$

□

Finally, note that proposition 4.2 allows us to calculate distances up to any prescribe degree of accuracy using iteration, provided the Kantorovich metrics can be efficiently and suitably calculated themselves. It remains to be seen if such a method will be feasible in practice.

## 5 Value Function Bounds

**Theorem 5.1.** *Suppose $\gamma \leq c$. Then $V^*$ is 1-Lipschitz continuous with respect to $d_{fix}^c$, i.e.*

$$|V^*(s) - V^*(s')| \leq d_{fix}^c(s, s').$$

*Proof.* Each iterate $V^n$ is continuous, and so each $|V^n(s) - V^n(s')|$ belongs to $\mathcal{M}$. The result now follows by induction and taking limits. □

## 6 Illustration

In this section we present a toy example of metric computation and metric approximation guarantees. Let $S = [0,1]$ with the usual Borel sigma-algebra, $A = \{a,b\}$, $r_s^a = 1-s$, $r_s^b = s$, $P_s^a$ be uniform on $S$, and $P_s^b$ the point mass at $s$. Clearly, these MDP parameters satisfy the required assumptions.

Given any $c \in (0,1)$, we claim

$$d_{fix}^c(x,y) = \frac{|x-y|}{1-c}.$$

Denote the RHS by $h$. Note that $T_K(h)(P_x^a, P_y^a) = 0$ and $T_K(h)(P_x^b, P_y^b) = \sup_{f \in Lip(h)} f(x) - f(y)$. Taking $f_1(x) = \frac{x}{1-c}$ and $f_2(x) = 1 - f_1(x)$ in $Lip(h)$ we find $T_K(h)(P_x^b, P_y^b) = h(x,y)$. Thus, $F^c(h)(x,y) = \max(|x-y| + c \cdot 0, |x-y| + c \cdot h(x,y)) = |x-y| + c \cdot h(x,y) = h(x,y)$. By uniqueness, $d_{fix}^c = h$.

Now consider the following approximation. Given $\epsilon > 0$, choose $n$ large enough so that $\frac{1}{n} < (1-c)\epsilon$. Partition $S$ as $B_k = [\frac{k}{n}, \frac{k+1}{n})$, $B_{n-1} = [\frac{n-1}{n}, 1]$, for $k = 0, 1, 2, \ldots, n-2$. Note that the diameter of each $B_k$ with respect to $d_{fix}^c$ is $\frac{1}{n(1-c)} < \epsilon$. The $n$ partitions will be the states of a finite MDP approximant. We obtain the rest of the parameters by averaging over the states in a partition. Thus, $r_{B_k}^a = 1 - \frac{2k+1}{2n}$, $r_{B_k}^b = \frac{2k+1}{2n}$, $P_{B_k,B_l}^a = \frac{1}{n}$, and $P_{B_k,B_l}^b = \delta_{B_k,B_l}$.

Assume $\gamma$ is given and choose $c = \gamma$. Note that for all $x, y \in B_k$, $|V^*(x) - V^*(y)| \leq diam_{d_{fix}^c} B_k \leq \epsilon$. Thus, we would expect that by averaging, and solving the finite MDP, $V^*(B_k)$ should differ by at most $\epsilon$ from $V^*(x)$, for any $x \in B_k$. In fact, in this case the value functions of the original MDP and of the finite approximant can be computed directly and we can verify this. For $x \in S$, $B_k$,

$$V^*(x) = \begin{cases} 1 - x + \frac{\gamma}{2(1-\gamma)} & \text{if } 0 \leq x < \frac{1}{2} \\ \frac{x}{1-\gamma} & \text{if } \frac{1}{2} \leq x \leq 1 \end{cases}$$

$$V^*(B_k) = \begin{cases} 1 - \frac{2k+1}{2n} + \frac{\gamma}{2(1-\gamma)} & \text{if } 0 \leq k < \frac{n-1}{2} \\ \frac{\frac{2k+1}{2n}}{1-\gamma} & \text{if } \frac{n-1}{2} \leq k \leq n-1 \end{cases}$$

Thus, for $x \in B_k$, $|V^*(x) - V^*(B_k)| \leq \frac{1}{1-\gamma}|x - \frac{2k+1}{2n}| \leq diam_{d_{fix}^c} B_k \leq \epsilon$.

# 7 Conclusion

In this paper we have constructed metrics for MDPs with continuous state spaces. Each metric has bisimulation as its kernel and is continuous in the MDP parameters. Most importantly, each metric bounds the optimal value of states continuously. Hence, if one was to aggregate states, this metric allows a guarantee on the error introduced by this approximation.

In contrast to previous situations, in this theoretical development the most important factor that we have to take into consideration was the way in which the rewards vary across the state space. What can be said in the case of a general bounded measurable, yet not necessarily continuous, reward function? In order to generalize our results, we need to establish the measurability of the map taking a pair of states to the Kantorovich distance, and to generalize lemma 3.1. We are currently working on this development. In the meantime, if the reward structure does not satisfy our assumption, we can still consider the best lsc approximations to $|r_s^a - r_{s'}^a|$ in $\mathcal{M}$. That is, we can replace $|r_s^a - r_{s'}^a|$ by $R_1^a(s,s') = \inf_{\mathcal{M}}\{|r_s^a - r_{s'}^a|\}$, and $R_2^a(s,s') = \sup_{\mathcal{M}}\{|r_s^a - r_{s'}^a|\}$ and obtain two fixed point semimetrics $d_1^c$ and $d_2^c$, respectively. Then it is not hard to modify theorem 4.1 to show that $Rel(d_2^c) \subseteq \sim \subseteq Rel(d_1^c)$. The idea is that we are sandwiching bisimulation by nearby closed equivalence relations.

The theoretical foundation we established can be used, potentially in two different ways. The first idea is to use the distance metric in the process of state aggregation, in order to provide a finite approximant for a continuous state MDP. However, even though our distance metrics are computable, the computation methods that we have investigated so far are not satisfactory. Discretizing the Kantorovich linear program may result in added complexity when one considers that the direct solution might be "simple". On the other hand, more direct methods of calculating the distance are not currently known in general.

The second idea, which holds a lot of promise, is to use our metric as a tool for the theoretical analysis of existing approximation schemes. There are many heuristic methods for providing variable resolution or multi-resolution approximations to MDPs with continuous state spaces. Using our metrics, the error bounds of these heuristics can be analyzed. A second important application is in the analysis of approximation schemes which start with a coarse approximant and gradually increase the resolution. The distance metrics can provide a tool for proving that such schemes converge to correct value estimates in the limit. We are currently pursuing research in this direction.

Recently, a promising Monte-Carlo based implementation of a finite approximation scheme for (continuous state space) LMPs has been developed (Bouchard et al., 2005). Aside from the practical importance of such an algorithm, it was shown that the finite approximants converged to the original system in the analogously defined LMP bisimulation metrics. Thus, one has the power of a realistic approximation scheme with theoretical guarantees. The current authors are more than hopeful that this work can easily be carried over to continuous state space MDPs. In that case, metric convergence along with value function bounds guarantee that optimal solutions to finite approximants are close to optimal solutions to the original system.

**Acknowledgments**

This work has been supported in part by funding from NSERC and CFI.

**Appendix**

**Lemma 7.1.** *Suppose $C$ is a closed equivalence relation on $S$. Then*

$$T_K(\mathbb{I}_{\mathscr{C}})(P,Q) = \sup_{X \in \Sigma(C)} |P(X) - Q(X)|.$$

*Proof.* For every $X \in \Sigma(C)$, the indicator function on $X$ belongs to $Lip(\mathbb{I}_{\mathscr{C}})$. Thus, the RHS is at most the LHS. For the other inequality, fix a positive $\epsilon$ and take $f : S \to [0,1]$ and $\psi = \sum_{i=1}^n c_i \cdot \mathbb{I}_{E_i}$ as in the proof of lemma 3.1. Let $J = \{i | P(E_i) \geq Q(E_i)\}$. Then

$$\begin{aligned}
T_K(\mathbb{I}_{\mathscr{C}})(P,Q) - 2\epsilon &\leq P(\psi) - (Q\psi) \\
&= \sum c_i(P(E_i) - Q(E_i)) \\
&\leq \sum_J c_i(P(E_i) - Q(E_i)) \\
&\leq (\max_J c_i) \sum_J (P(E_i) - Q(E_i)) \\
&\leq 1 \cdot (P(\cup_J E_i) - Q(\cup_J E_i)) \\
&\leq \sup_{X \in \Sigma(C)} |P(X) - Q(X)|
\end{aligned}$$

since $\cup_J E_i$ belongs to $\Sigma(C)$. □